\pdfoutput=1
\documentclass[runningheads]{llncs}
\usepackage{url}
\usepackage{subfig}
\usepackage{booktabs}
\usepackage{graphicx}
\usepackage[colorlinks=true,urlcolor= blue]{hyperref}%
\usepackage{bbding}

\begin{document}
\title{Multi-Task Learning for Left Atrial Segmentation on GE-MRI}
%
%
\author{A. Anonymous}
\author{Chen Chen\inst{1}(\Envelope)\and
Wenjia Bai \inst{1}\and
Daniel Rueckert \inst{1}}
\authorrunning{Chen. et al.}
%
\institute{Biomedical Image Analysis Group, Imperial College London, London, UK\\
\email{cc215@ic.ac.uk}}
\maketitle              
\begin{abstract}
Segmentation of the left atrium (LA) is crucial for assessing its anatomy in both pre-operative atrial fibrillation (AF) ablation planning and post-operative follow-up studies. In this paper, we present a fully automated framework for left atrial segmentation in gadolinium-enhanced magnetic resonance images (GE-MRI) based on deep learning. We propose a fully convolutional neural network and explore the benefits of multi-task learning for performing both atrial segmentation and pre/post ablation classification. Our results show that, by sharing features between related tasks, the network can gain additional anatomical information and achieve more accurate atrial segmentation, leading to a mean Dice score of 0.901 on a test set of 20 3D MRI images. Code of our proposed algorithm is available at~\url{https://github.com/cherise215/atria_segmentation_2018/}.
\end{abstract}
\section{Introduction}
Atrial fibrillation (AF) is a condition of the heart that causes an irregular and often abnormally fast heart rate~\cite{nhs_choices_2018}. This can cause blood clots to form, which can restrict blood supply to vital organs, and further leads to a stroke and heart failure~\cite{calkins20122012}.  One of the most common treatments for AF is called ablation which can isolate the pulmonary veins (PVs) from the Left Atrium (LA) electrically by inducing circumferential lesion and destroying abnormal tissues. During this procedure, a good understanding of the patient atrial anatomy is very vital for planning and guiding the surgery, and further improving the patient outcome~\cite{calkins20122012}. 

A good way to learn the anatomical structure of the LA is by performing  LA segmentation on medical images, such as computed tomography (CT) scans and magnetic resonance images (MRI). With the development of imaging techniques and computer science, many automatic or semi-automatic algorithms~\cite{tobon2015benchmark} have been proposed for atrial segmentation. However, this is still a challenging problem and many traditional methods may fail to segment due to several reasons. For example, intensity-based methods such as region growing may fail to segment those atria with extremely thin myocardial walls, especially when their surroundings have very similar intensity to their blood pool~\cite{tobon2015benchmark}. In addition, there is large shape variation among the LA of different individuals, such as atrial sizes and pulmonary vein structures~\cite{tobon2015benchmark}. These variations will make it too complex for model-based segmentation methods to impose shape prior. An alternative way is to use atlas-based methods that can be robust to the LA with high anatomical variations. However, this kind of approach is time-consuming which typically takes 8 minutes around~\cite{Depa2010}. Most recently, with the increase of computing hardware performance and more data becoming available, deep learning has become the state-of-the-art method due to its efficiency and effectiveness on computer vision tasks, and has been widely used in the medical domain~\cite{litjens2017survey}.

In this paper, we focus on the segmentation of the LA from gadolinium-enhanced magnetic
resonance images (GE-MRI). These images can either be taken before or after ablation treatment. Noticing that there might be contextual difference between the pre-ablation and post-ablation images (e.g. ablation will cause scars in the LA~\cite{Malcolme2013} and may influence the quality of images), we proposed a multi-task convolutional neural network (CNN)
that could segment a patient left atrium from GE-MRI and detect whether this patient is pre- or
post-ablation. In this way, our network could not only learn structural information from segmentation masks, but also retrieve contextual information through the classification task. Our network was trained simultaneously for the two tasks, using a stack of 2D slices extracted from each MRI scan along with its corresponding segmentation masks and a pre/post ablation label. In addition, in order to improve the robustness of segmentation on images with various image contrast and sizes, we employed a contrast augmentation method to augment our training set and trained our network with images in different sizes. In order to produce a fixed-length vector to classify input images in multiple sizes, spatial pyramid pooling~\cite{he2014spatial} was used in this network. 

The proposed framework was trained and evaluated on the data set of the Atrial Segmentation Challenge 2018. 
Our experimental results showed that by sharing features between related tasks, our network achieved better segmentation performance compared to a variant of U-Net trained with a single task. During the test phase, our network can directly inference the segmentation mask from a scan of GE-MRI without taking extra pre-processing steps for image contrast enhancement. In total, our method is very efficient as one 3D segmentation result for each individual was obtained in 6 seconds on a Nvidia Titan Xp GPU using our model, plus 3 or 4 seconds for post-processing on the whole volume, which is far more faster than general atlas-based methods that usually take minutes.

\section{Methods}
In this section, we present the architecture of our proposed multi-task network and how we post-process the network output to get the final 3D segmentation mask. Our proposed network is adapted from a U-Net architecture~\cite{RonnebergerFB15} where we increase the depth of the network and add a classification branch. The input to the network is a stack of 2D images. The output is a one shot prediction of the atrial segmentation mask and pre/post ablation classification score. 

\subsection{Network Architectures}
In order to explore the benefit of multi-task learning, the proposed network is designed to conduct both the atrial segmentation task and an auxiliary pre/post ablation classification task with images of multiple sizes. 

\begin{figure}[]
\centering
\includegraphics[width=\textwidth]{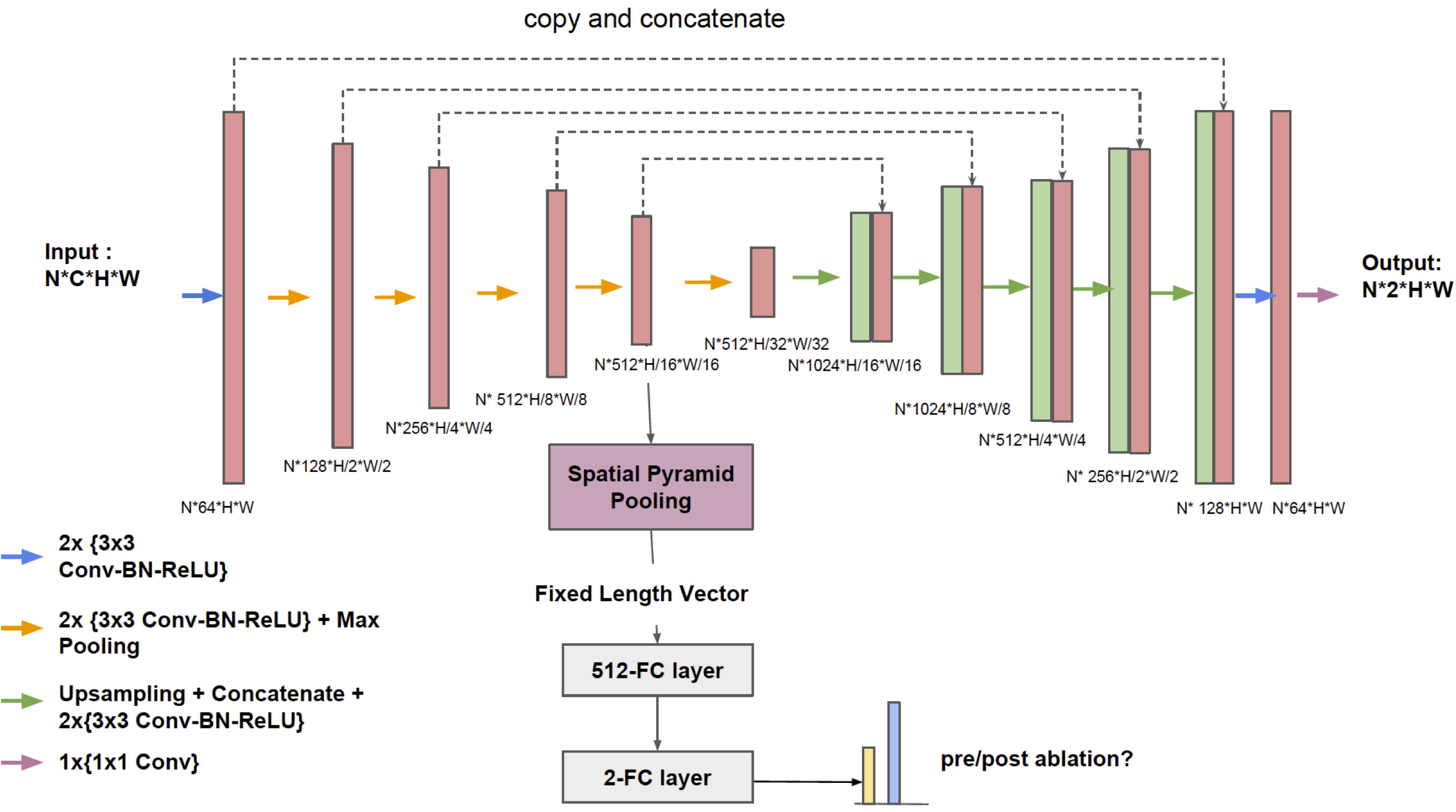}
\caption{The Architecture of our proposed multi-task Deep U-Net. Best viewed in color}
\label{network structure}
\end{figure}

The core of our method, named `Deep U-Net' and shown in Fig~\ref{network structure}, is derived from the 2D U-Net~\cite{RonnebergerFB15} for semantic segmentation. Since the largest size of images in our dataset is $640\times640$ in x-y planes, we increased the receptive field of U-Net by adding more pooling layers. The modified network now consists of five down-sampling blocks and five up-sampling blocks. Each down-sampling layer contains two 3x3 convolutions, with Batch Normalizations(BN)~\cite{IoffeS15} and Rectified Linear Unit (ReLU) activations, as well as a 2x2 max pooling operation with stride 2 for down-sampling. The up-sampling path is symmetric to the down-sampling path. By aggregating both coarse and fine features learned at different scales from the down-sampling path and up-sampling path, our network is supposed to achieve better segmentation performance than those networks without the aggregation operations.

Our classification task is performed by utilizing image features learned from the down-sampling path. Features after the $4^{th}$ max pooling layer are extracted for classification, which is a common practice for many existing classification networks~\cite{Krizhevsky2012,2015arXiv151203385H}. In order to generate fix-length feature vectors learned from input images with different sizes and scales, spatial pyramid pooling~\cite{he2014spatial} is applied. These fixed-length vectors are then processed through Fully-Connected layers (FC) followed by a softmax layer to calculate class probabilities (pre/post-ablation) for each image. Dropout~\cite{srivastava2014dropout} is applied to the output of FC layers with a probability of 0.5 during the training process, which functions as regularizer to encourage our network to better generalize~\cite{srivastava2014dropout}. 

The loss function $L$ for our multi-task network is $L=L_S+\lambda L_C$, where $L_S$ is segmentation score, $L_C$ is classification score, and $\lambda = 1$ for our experiments. For segmentation part, pixel-wise cross-entropy loss was employed. For classification part, we used sigmoid cross-entropy to measure pre/post ablation classification loss. The classification ground truth of every 2D image is labeled as one if this slice is extracted from a post-ablation object. Otherwise, its ground truth is zero. Our network was trained jointly with the combined loss. The classification loss works as a regularization term, enabling the network to learn the high-level representation that generalizes well on both tasks. 

Our network was optimized using Stochastic Gradient Descent(SGD)~\cite{rumelhart1986learning} with a momentum of $0.99$ and weight decay of $0.0005$. The initial learning rate is 0.001, which will be decreased at a rate of 0.5 after every 50 epochs.

\subsection{Post-processing}
During the inference time, axial slices extracted from a 3D image are fed into the network slice by slice. The segmentation branch predicts pixel-wise probability score for both background and atrium classes. A 2D segmentation mask is then generated by finding the class with the highest probability for each pixel on the slice. By concatenating these segmentation results slice by slice,a rough 3D mask for each patient is produced. In order to refine the boundary of those masks, we performed 3D morphological dilation and erosion, and kept the largest connected component for each volume.

\section{Experiments and Results}
\subsection{Data}
In this paper, our algorithm was trained and evaluated on the dataset of the 2018 Atrial Segmentation Challenge~\footnote{\url{http://atriaseg2018.cardiacatlas.org/}}.
This dataset contains a training set of 100 3D gadolinium-enhanced magnetic resonance imaging scans (GE-MRIs) along with corresponding LA manual segmentation mask and pre/post ablation labels for training and validation. In addition, there is a set of 54 images without labels provided for testing. For model training and evaluation, we randomly split the training set into $80:20$. We did not use any external data for training or pre-training of our network.

\begin{figure}[!ht]
  \centering
  \subfloat[Patient A]{
    \includegraphics[height=3cm]{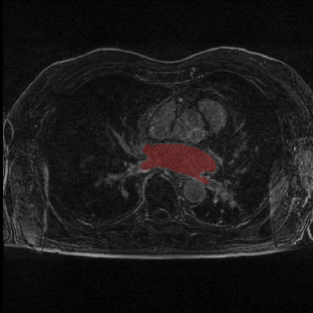} \hfill
      \includegraphics[height=3cm]{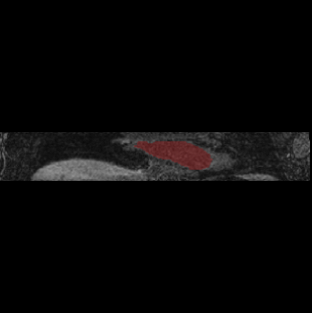} \hfill
       \includegraphics[height=3cm]{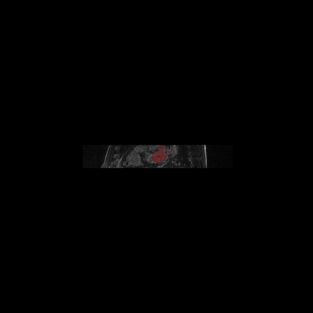} \hfill
  } \hfill
  \subfloat[Patient B]{
    \includegraphics[height=3cm]{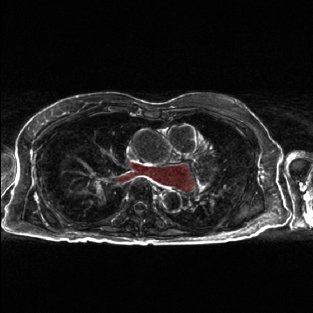} \hfill
    \includegraphics[height=3cm]{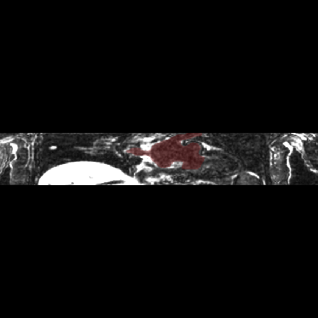} \hfill
      \includegraphics[height=3cm]{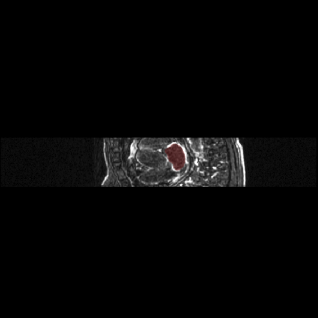} \hfill
  } \hfill
  \caption{Original images slices from different views. Despite the homogeneity in image resolution, there are significant differences in the image contrast and quality among different individuals, which can be challenging to segment the Left Atrium (red) from MRI images. Best viewed in color}
  \label{data_visualize}
\end{figure}

Images in this dataset have been resampled and preprocessed by the organizers. So there is no need to do re-sampling procedure in the pre-processing stage. Despite the consistency observed in the resolution of the data, this dataset exhibits large differences in images sizes and image contrast. For example, in our MRI dataset, there are two sizes of images: $576\times576$ and $640\times640$ on the axial planes. Apart from that, atria, in different images, can also have various shapes and sizes. These phenomena may arise due to the fact that these scans were collected from multiple sites which may have different scanners and imaging protocols. Hence, it is important to build a robust method for those images. Fig.~\ref{data_visualize} visualizes the difference in image contrast of different images in different views. In this paper, we use data augmentation to increase data variety with the aim of improving the model's generalization ability on different images, which will be discussed in section~\ref{dataaug}. 

\subsection{Data Pre-processing}
In order to preserve the resolution of images, image re-scaling was not performed in the data pre-processing stage. Instead, multi-scale cropping was used to increase the data variety, so that network can analyze images with different contexts. More details will be described in the next section as it actually happens in the data augmentation process. For testing, images can be directly fed into the network provided that its length is a factor of 32 due to the architecture of the network. Otherwise, reflect padding is required. The only necessary step in our pre-processing stage in both training and testing stage is to normalize image intensity to zero mean and unit variance, which has been widely accepted in common practice.

\subsection{Data Augmentation}
\label{dataaug}
Our training data was augmented via a composition of image transformations,  including random horizontal/vertical flip with a probability of $50\%$, random rotation with degree range from $-10$ to $+10$, random shifting along X and Y axis within the range of 10 percent of its original image size, and zooming with a factor between 0.7 and 1.3. 

In addition, we also employed random gamma correction as a way of contrast augmentation. Image contrast was adjusted based on a point-wise nonlinear transformation: $G(x,y)=F(x,y)^{1/\gamma}$ where $F(x,y)$ is the original value of each pixel in an image, and $G(x,y)$ is the transformed value for each pixel.  The value of $\gamma$ is randomly chosen from the range of $(0.8,2.0)$ for each image. By applying gamma correction randomly, the variety of image contrast in the training set was significantly increased. 

In order to process images at multiple sizes with objects at multiple scales, we centrally cropped 2D images at various image scales. The cropped sizes include $256\times256,384\times384,480\times480,512\times512,576\times576,640\times640$. If the cropped size was larger than the image's original size, mirror padding was performed instead. Motivated by Curriculum Learning~\cite{bengio2009curriculum}, we trained our network firstly with cropped images where the left atrium taking a large portion of the image and then we gradually increased the image size. In this way, our network learns to segment from easy scenarios to hard scenarios and this helps the model to quickly converge in the beginning~\cite{Lieman-Sifry2017}. Despite the change in input images sizes, our network could still output a fixed length feature vector for classification since we employed spatial pyramid pooling \cite{he2014spatial}. In practice, we found this could help the network focus on learning task-specific structural features for organ segmentation regardless of the contextual changes in sizes and scales. It is also beneficial for quantitative analysis based on medical image segmentation since we do not use rescaling nor resizing operations which have the risk of introducing scaling/shifting artifacts during prediction. \\

\noindent\textbf{Contrast Augmentation:}
We found that there exists a diversity of image contrast in the dataset, where low-contrast effects can reduce the visual quality of an image~\cite{Somasundaram2011} and thus affect segmentation accuracy. To solve this issue, traditional machine learning methods often require image contrast enhancement methods during image pre-processing. Here, we proposed a contrast augmentation method based on gamma correction instead, to generate a variety of images with different levels of contrast during training. In this way, our CNN could gain the ability to segment images regardless of the difference of image contrast. And there is no need to do any contrast adjustment during testing. Therefore, our method is more efficient than those general traditional methods which require those adjustments. 

To show our contrast augmentation method is superior to the traditional contrast enhancement methods, we compared it with two image contrast enhancement methods: contrast limited adaptive histogram equalization (CLAHE)~\cite{Zuiderveld1994} and automatic gamma correction~\cite{Somasundaram2011}. Both of them have been widely used in the pre-processing of CT image and MRI image applications~\cite{pizer1990contrast,selvy2013proficient,ghose2012random} in order to improve medical image quality for visual tasks. For CLAHE, we divided each image into $8\times8$ regions and performed contrast enhancement on each region by default.

The above experiments were performed based on a simple Deep U-Net (without multi-task) for comparison. From Table~\ref{tab1}, it can be seen that our proposed data augmentation method could significantly improve the robustness of our network for processing images with various image contrast and outperformed the traditional image pre-processing methods which may have the risk of amplifying noises and take extra processing time. Therefore, in the following sections, we would like to employ contrast augmentation as our default experimental setting.

\label{contrast augmentation}
\begin{table}[!ht]
\centering
\caption{Segmentation accuracy using a single-task Deep U-net with different contrast processing strategies}
\label{tab1}
\begin{tabular}{@{}cccc@{}}
\toprule
Base Model & Method                     & Need Extra Time & Dice          \\ \midrule
Deep U-Net  & Standard Normalization     & No              & 0.847 (0.18) \\
Deep U-Net  & Automatic Gamma Correction & Yes             & 0.854 (0.15) \\
Deep U-Net  & CLAHE                      & Yes             & 0.876 (0.09) \\
Deep U-Net  & Gamma Augmentation         & No              & 0.883 (0.08) \\ \bottomrule
\end{tabular}
\end{table}

\subsection{Results}
To evaluate our segmentation accuracy for different experimental settings, we use four measurements: the Dice score (also known as Dice similarity coefficient score), the Jaccard Similarity Coefficient (JC) score, the Hausdorff Distance (HD) and the Average Symmetric Surface Distance (ASSD).

\begin{table}[!ht]
\centering
\caption{Segmentation accuracy results based on different measurements for different networks and methods}
\label{comparison of U-Net with our networks}
\begin{tabular}{@{}ccccc@{}}
\toprule
                                                                                      & Dice                  & JC                     & HD                   & ASSD                  \\
\midrule
Vanilla U-Net                                                                         & 0.855 (0.11)       & 0.760 (0.14)          & 21.81 (19.35)         & 1.577 (1.07)          \\
Deep U-Net                                                                            & 0.883 (0.08)        & 0.798 (0.11)           & 21.18 (21.00)         & 1.199 (0.47)          \\
Deep U-Net + multi-task
                                                             & 0.896 (0.04)          & 0.815 (0.07)           & 15.40 (6.39)          & 1.11 (0.35)          \\
\begin{tabular}[c]{@{}c@{}}Deep U-Net + multi-task\\+ post-proc. \end{tabular} & ~\textbf{0.901 (0.03)} & \textbf{0.822 (0.06)} & \textbf{14.23 (4.83)} & \textbf{1.04 (0.32)}   \\
\bottomrule
\end{tabular}
\end{table}

\begin{figure}[!ht]
\includegraphics[width=\textwidth]{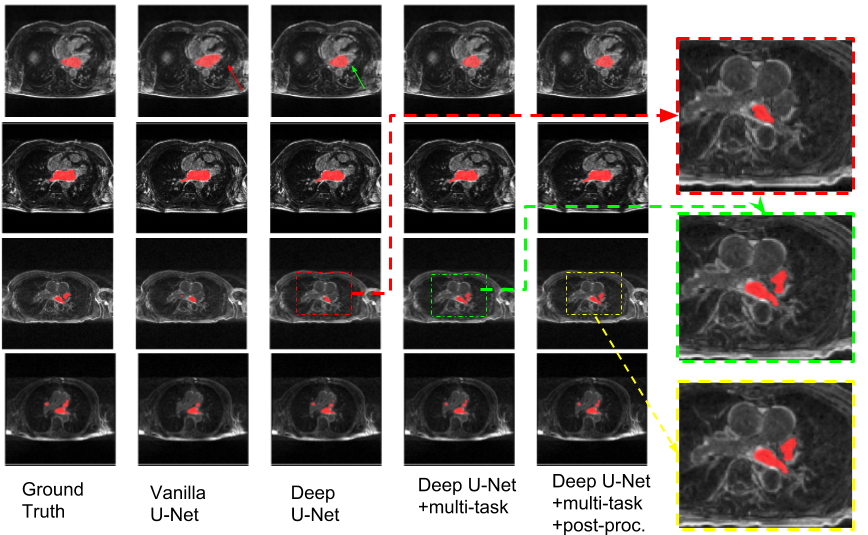}
\caption{Example segmentations for axial slices from our MRI set using different methods. Each column shows axial slices from the mitral to the PVs plane from different individuals (top to bottom).}
\label{fig:Result_of_Atrial_Seg}
\end{figure}

\begin{figure}[!ht]
  \centering
{
    \includegraphics[height=1.8cm]{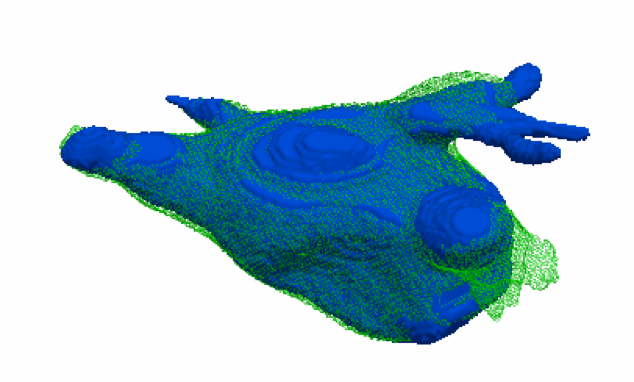} 
  } \hfill
 {
    \includegraphics[height=1.8cm]{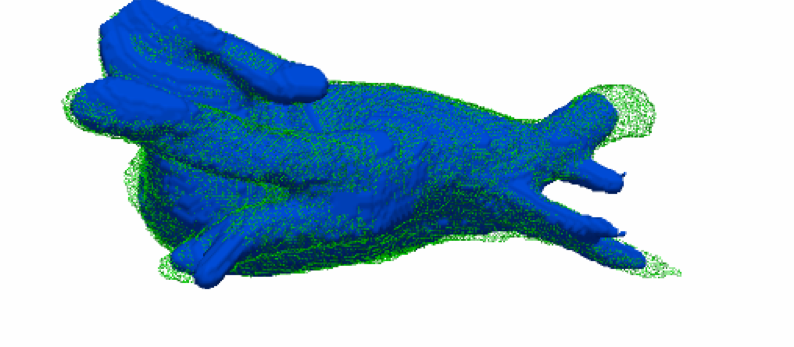} 
  } \hfill
 {
    \includegraphics[height=1.8cm]{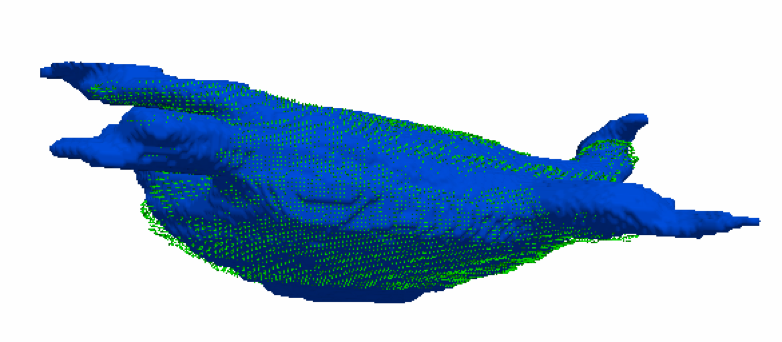} 
  } \hfill
  \caption{3D visualization of three samples from the validation set: blue objects are the ground truth, and the green ones are the predicted segmentation of our proposed method.}
  \label{3D_visualize}
\end{figure}

To show the advancement of our deep network with additional pooling/max-pooling layers, we compared our modified networks with the vanilla 2D U-Net. The results are shown in Table~\ref{comparison of U-Net with our networks}. It can be seen that the segmentation performance was greatly improved by increasing the depth of the network, especially in Mitral Valve(MV) planes. Our best results were achieved by using the multi-task Deep U-Net followed by post-processing, producing a Dice score of 0.901. From the visualization plots in Fig.~\ref{fig:Result_of_Atrial_Seg}, we could see that our multi-task U-Net is more robust than the other two with only one segmentation goal. One reason could be that by sharing features with segmentation and related pre/post ablation classification, the network is forced to learn better representation on images taken before the ablation treatment and those after the treatment, which could further improve segmentation performance. Fig.~\ref{3D_visualize} showed that our model achieved high overlap ratio between our 3D segmentation result and the ground truth in different subjects. However, one significant failure mode can be observed around the region of pulmonary veins. One possible reason might be that the number and the length of pulmonary veins vary from person to person, making it too hard for the network to learn from limited cases.

The total processing procedure (inference + post-processing) for each whole 3D MRI predicted by our network took approximately 10 seconds on average on one Nvidia Titan Xp GPU. It is therefore much more efficient than those atlas-based methods which typically take eight minutes~\cite{Depa2010}. 

For the Atrial Segmentation Challenge 2018, we adopted an ensemble method called Boostrap Aggregating (Bagging)~\cite{breiman1996bagging} to improve our model's performance in the test phase. We noticed that samples in the dataset were collected from multiple sites while a large portion is from The University of Utah. In that case, domain shift or domain bias may exist when we use a model trained on one certain subset from a limited dataset to predict data from another subset as they may have different intensity distributions. Therefore, we trained the same model 5 times, each with a random subset and then averaged the class probabilities produced by these five models for prediction. Our ensembled results on a set of 54 test cases given by the organizers improved from an averaged Dice score of 0.9197 to 0.9206.

\section{Conclusion}
In this paper, we proposed a deep 2D fully convolutional neural network to automatically segment the left atrium from GE-MRIs. By applying multi-task learning, our network demonstrated improved segmentation accuracy compared to a baseline U-Net method. In addition, we showed that contrast augmentation is an efficient and effective way to enhance our model's robustness and efficiency when analyzing images with various image contrast. 
\bibliographystyle{unsrt}
\bibliography{mybib}
\end{document}